\documentclass[11pt,a4paper]{article}
\usepackage[hyperref]{acl2021}
\usepackage{times}
\usepackage{latexsym}

\usepackage{microtype}

\aclfinalcopy 

\title{Factoring Statutory Reasoning as Language Understanding Challenges}

\author{Nils Holzenberger \and Benjamin Van Durme \\
  Center for Language and Speech Processing \\
  Johns Hopkins University \\
  Baltimore, Maryland, USA \\
  \texttt{nilsh@jhu.edu} $\quad$ \texttt{vandurme@cs.jhu.edu}}

\date{}

\usepackage{graphicx,wrapfig,bm}
\newcommand{\mysubsubsection}[1]{\paragraph{#1}}
\newcommand{\myvector}[1]{\boldsymbol{#1}}
\newcommand{\mymathcomma}[0]{,\allowbreak}
\newcommand{\myexample}[1]{\vspace{5pt} \noindent \textbf{#1} \vspace{5pt}}
\usepackage{algorithm,algorithmicx,algpseudocode}
\usepackage{colortbl}
\usepackage{enumitem}
\usepackage{needspace}
\usepackage{amsfonts}
\usepackage[all]{nowidow}

\begin{document}
\maketitle
\begin{abstract}
Statutory reasoning is the task of determining whether a legal statute, stated in natural language, applies to the text description of a case. Prior work introduced a resource that approached statutory reasoning as a monolithic textual entailment problem, with neural baselines performing nearly at-chance. To address this challenge, we decompose statutory reasoning into four types of language-understanding challenge problems, through the introduction of concepts and structure found in Prolog programs. Augmenting an existing benchmark, we provide annotations for the four tasks, and baselines for three of them. Models for statutory reasoning are shown to benefit from the additional structure, improving on prior baselines. Further, the decomposition into subtasks facilitates finer-grained model diagnostics and clearer incremental progress.
\end{abstract}

\section{Introduction}

As more data becomes available, Natural Language Processing (NLP) techniques are increasingly being applied to the legal domain, including for the prediction of case outcomes \citep{xiao2018cail, vacek-etal-2019-litigation,chalkidis-etal-2019-neural}. In the US, cases are decided based on previous case outcomes, but also on the legal statutes compiled in the US code. For our purposes, a \emph{case} is a set of facts described in natural language, as in Figure~\ref{fig:subtasks}, in blue. The US code is a set of documents called \emph{statutes}, themselves decomposed into \emph{subsections}. Taken together, subsections can be viewed as a body of interdependent rules specified in natural language, prescribing how case outcomes are to be determined. \emph{Statutory reasoning} is the task of determining whether a given subsection of a statute applies to a given case, where both are expressed in natural language. Subsections are implicitly framed as predicates, which may be true or false of a given case. \citet{holzenberger2020dataset} introduced SARA, a benchmark for the task of statutory reasoning, as well as two different approaches to solving this problem. First, a manually-crafted symbolic reasoner based on Prolog is shown to perfectly solve the task, at the expense of experts writing the Prolog code and translating the natural language case descriptions into Prolog-understandable facts. The second approach is based on statistical machine learning models. While these models can be induced computationally, they perform poorly because the complexity of the task far surpasses the amount of training data available.

We posit that statutory reasoning as presented to statistical models is underspecified, in that it was cast as Recognizing Textual Entailment \citep{dagan2005pascal} and linear regression. Taking inspiration from the structure of Prolog programs, we re-frame statutory reasoning as a sequence of four tasks, prompting us to introduce a novel extension of the SARA dataset (Section~\ref{sec:sara}), referred to as \emph{SARA v2}. Beyond improving the model's performance, as shown in Section~\ref{sec:baseline}, the additional structure makes it more interpretable, and so more suitable for practical applications. We put our results in perspective in Section~\ref{sec:discussion} and review related work in Section~\ref{sec:related_work}.

\section{SARA v2}
\label{sec:sara}

\begin{figure*}[h!]
    \centering
    \includegraphics[width=\textwidth]{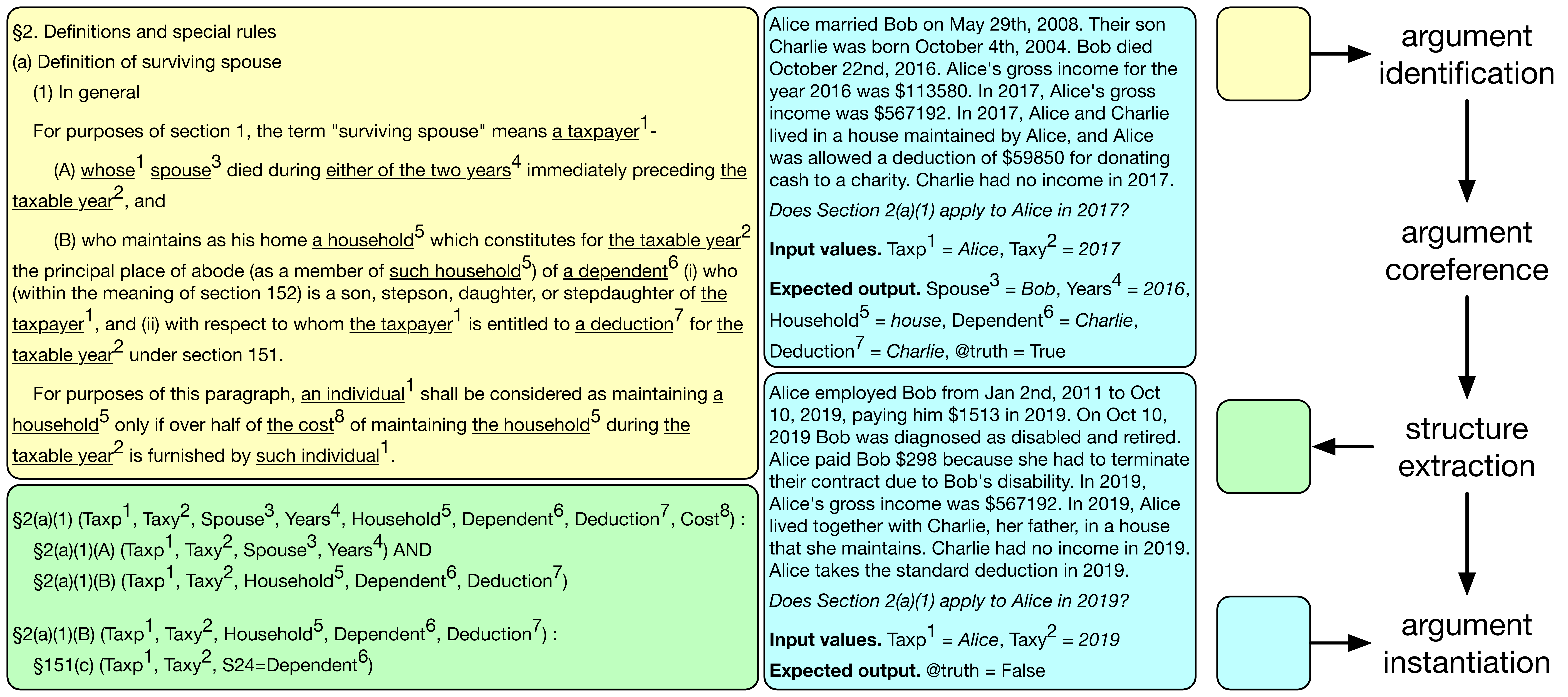}
    \caption{Decomposing statutory reasoning into four tasks. The flowchart on the right indicates the ordering, inputs and outputs of the tasks. In the statutes in the yellow box, argument placeholders are underlined, and superscripts indicate argument coreference. The green box shows the logical structure of the statutes just above it. In blue are two examples of argument instantiation.}
    \label{fig:subtasks}
\end{figure*}

The symbolic solver requires experts translating the statutes and each new case's description into Prolog. In contrast, a machine learning-based model has the potential to generalize to unseen cases and to changing legislation, a significant advantage for a practical application. In the following, we argue that legal statutes share features with the symbolic solver's first-order logic. We formalize this connection in a series of four challenge tasks, described in this section, and depicted in Figure~\ref{fig:subtasks}. We hope they provide structure to the problem, and a more efficient inductive bias for machine learning algorithms. The annotations mentioned throughout the remainder of this section were developed by the authors, entirely by hand, with regular guidance from a legal scholar\footnote{The dataset can be found under \url{https://nlp.jhu.edu/law/}}. Examples for each task are given in Appendix~\ref{app:task_examples}. Statistics are shown in Figure~\ref{fig:stats} and further detailed in Appendix~\ref{app:dataset_statistics}.

\paragraph{Argument identification}

This first task, in conjunction with the second, aims to identify the arguments of the predicate that a given subsection represents. Some terms in a subsection refer to something concrete, such as ``the United States'' or ``April 24th, 2017''. Other terms can take a range of values depending on the case at hand, and act as placeholders. For example, in the top left box of Figure~\ref{fig:subtasks}, the terms ``a taxpayer'' and ``the taxable year'' can take different values based on the context, while the terms ``section 152'' and ``this paragraph'' have concrete, immutable values. Formally, given a sequence of tokens ${t_1, ..., t_n}$, the task is to return a set of start and end indices ${(s,e) \in \{1, 2, ..., n\}^2}$ where each pair represents a span. We borrow from the terminology of predicate argument alignment \citep{roth2012aligning,wolfe-etal-2013-parma} and call these placeholders \emph{arguments}. The first task, which we call \emph{argument identification}, is tagging which parts of a subsection denote such placeholders. We provide annotations for argument identification as character-level spans representing arguments. Since each span is a pointer to the corresponding argument, we made each span the shortest meaningful phrase. Figure~\ref{fig:stats}(b) shows corpus statistics about placeholders.

\paragraph{Argument coreference}

Some arguments detected in the previous task may appear multiple times within the same subsection. For instance, in the top left of Figure~\ref{fig:subtasks}, the variable representing the taxpayer in §2(a)(1)(B) is referred to twice. We refer to the task of resolving this coreference problem at the level of the subsection as \emph{argument coreference}. While this coreference can span across subsections, as is the case in Figure~\ref{fig:subtasks}, we intentionally leave it to the next task. Keeping the notation of the above paragraph, given a set of spans ${\{(s_i,e_i)\}_{i=1}^S}$, the task is to return a matrix ${C\in \{0,1\}^{S \times S}}$ where ${C_{i,j} = 1}$ if spans ${(s_i,e_i)}$ and ${(s_j,e_j)}$ denote the same variable, $0$ otherwise. Corpus statistics about argument coreference can be found in Figure~\ref{fig:stats}(a). After these first two tasks, we can extract a set of arguments for every subsection. In Figure~\ref{fig:subtasks}, for §2(a)(1)(A), that would be \{\texttt{Taxp}, \texttt{Taxy}, \texttt{Spouse}, \texttt{Years}\}, as shown in the bottom left of Figure~\ref{fig:subtasks}.

\begin{figure*}[h!]
    \centering
    \includegraphics[width=\textwidth]{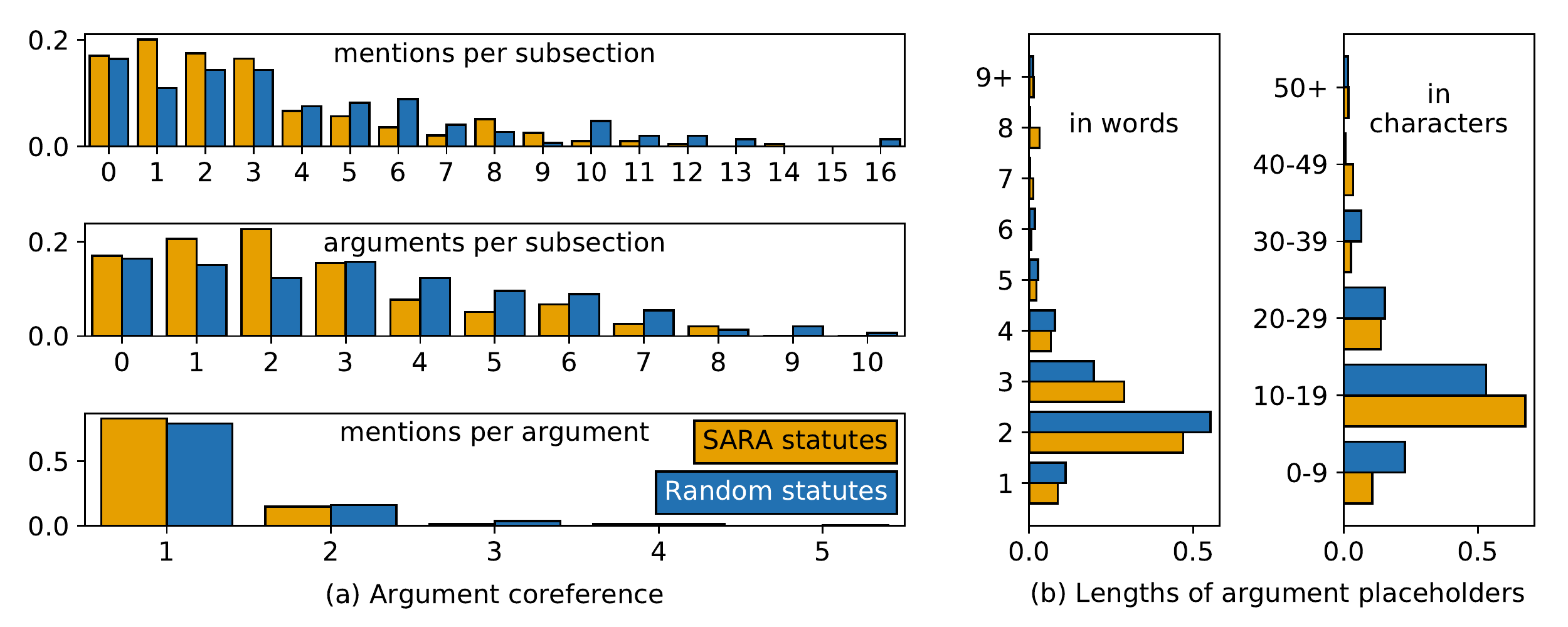}
    \caption{Corpus statistics about arguments. ``Random statutes'' are 9 sections sampled from the US code.}
    \vspace{-1pt}
    \label{fig:stats}
\end{figure*}

\paragraph{Structure extraction}

A prominent feature of legal statutes is the presence of references, implicit and explicit, to other parts of the statutes. Resolving references and their logical connections, and passing arguments appropriately from one subsection to the other, are major steps in statutory reasoning. We refer to this as \emph{structure extraction}. This mapping can be trivial, with the taxpayer and taxable year generally staying the same across subsections. Some mappings are more involved, such as the taxpayer from §152(b)(1) becoming the dependent in §152(a). Providing annotations for this task in general requires expert knowledge, as many references are implicit, and some must be resolved using guidance from Treasury Regulations. Our approach contrasts with recent efforts in breaking down complex questions into atomic questions, with the possibility of referring to previous answers \citep{wolfson-etal-2020-break}. Statutes contain their own breakdown into atomic questions. In addition, our structure is interpretable by a Prolog engine.

We provide structure extraction annotations for SARA in the style of Horn clauses \citep{horn1951}, using common logical operators, as shown in the bottom left of Figure~\ref{fig:subtasks}. We also provide character offsets for the start and end of each  subsection. Argument identification and coreference, and structure extraction can be done with the statutes only. They correspond to extracting a shallow version of the symbolic solver of \citet{holzenberger2020dataset}.

\paragraph{Argument instantiation}

We frame legal statutes as a set of predicates specified in natural language. Each subsection has a number of arguments, provided by the preceding tasks. Given the description of a case, each argument may or may not be associated with a value. Each subsection has an \texttt{@truth} argument, with possible values \emph{True} or \emph{False}, reflecting whether the subsection applies or not. Concretely, the input is (1) the string representation of the subsection, (2) the annotations from the first three tasks, and (3) values for some or all of its arguments. Arguments and values are represented as an array of key-value pairs, where the names of arguments specified in the structure annotations are used as keys. In Figure~\ref{fig:subtasks}, compare the names of arguments in the green box with the key names in the blue boxes. The output is values for its arguments, in particular for the \texttt{@truth} argument. In the example of the top right in Figure~\ref{fig:subtasks}, the input values are \texttt{taxpayer}~=~\emph{Alice} and \texttt{taxable year}~=~\emph{2017}, and one expected output is \texttt{@truth}~=~True. We refer to this task as \emph{argument instantiation}. Values for arguments can be found as spans in the case description, or must be predicted based on the case description. The latter happens often for dollar amounts, where incomes must be added, or tax must be computed. Figure~\ref{fig:subtasks} shows two examples of this task, in blue.

Before determining whether a subsection applies, it may be necessary to infer the values of unspecified arguments. For example, in the top of Figure~\ref{fig:subtasks}, it is necessary to determine who Alice's deceased spouse and who the dependent mentioned in §2(a)(1)(B) are. If applicable, we provide values for these arguments, not as inputs, but as additional supervision for the model. We provide manual annotations for all (subsection, case) pairs in SARA. In addition, we run the Prolog solver of \citet{holzenberger2020dataset} to generate annotations for all possible (subsection, case) pairs, to be used as a \emph{silver} standard, in contrast to the \emph{gold} manual annotations. We exclude from the silver data any (subsection, case) pair where the case is part of the test set. This increases the amount of available training data by a factor of 210.

\section{Baseline models}
\label{sec:baseline}

We provide baselines for three tasks, omitting structure extraction because it is the one task with the highest return on human annotation effort\footnote{Code for the experiments can be found under \url{https://github.com/SgfdDttt/sara_v2}}. In other words, if humans could annotate for any of these four tasks, structure extraction is where we posit their involvement would be the most worthwhile. Further, \citet{pertierra2017towards} have shown that the related task of semantic parsing of legal statutes is a difficult task, calling for a complex model.

\subsection{Argument identification}

We run the Stanford parser \citep{socher2013parsing} on the statutes, and extract all noun phrases as spans~-- specifically, all NNP, NNPS, PRP\$, NP and NML constituents. While de-formatting legal text can boost parser performance \citep{morgenstern2014toward}, we found it made little difference in our case.

As an orthogonal approach, we train a BERT-based CRF model for the task of BIO tagging. With the 9 sections in the SARA v2 statutes, we create 7 equally-sized splits by grouping §68, 3301 and 7703 into a single split. We run a 7-fold cross-validation, using 1 split as a dev set, 1 split as a test set, and the remaining as training data. We embed each paragraph using BERT, classify each contextual subword embedding into a 3-dimensional logit with a linear layer, and run a CRF \citep{lafferty2001conditional}. The model is trained with gradient descent to maximize the log-likelihood of the sequence of gold tags. We experiment with using Legal BERT \citep{holzenberger2020dataset} and BERT-base-cased \citep{devlin2018bert} as our BERT model. We freeze its parameters and optionally unfreeze the last layer. We use a batch size of 32 paragraphs, a learning rate of $10^{-3}$ and the Adam optimizer \citep{kingma2014adam}. Based on F1 score measured on the dev set, the best model uses Legal BERT and unfreezes its last layer. Test results are shown in Table~\ref{tab:argument_identification_results}.

\begin{table}[h!]
    \centering
    \begin{tabular}{lcc}
        \textbf{Parser-based}       & avg $\pm$ stddev & macro \\
        \hline precision            & 17.6 $\pm$ 4.4   & 16.6 \\
        recall                      & 77.9 $\pm$ 5.0   & 77.3 \\
        F1                          & 28.6 $\pm$ 6.2   & 27.3 \\
        \textbf{BERT-based} & avg $\pm$ stddev & macro \\
        \hline precision            & 64.7 $\pm$ 15.0  & 65.1 \\
        recall                      & 69.0 $\pm$ 24.2  & 59.8 \\
        F1                          & 66.2 $\pm$ 20.5  & 62.4 \\
    \end{tabular}
    \caption{Argument identification results. Average and standard deviations are computed across test splits.}
    \label{tab:argument_identification_results}
\end{table}

\subsection{Argument coreference}

Argument coreference differs from the usual coreference task \citep{pradhan-EtAl:2014:P14-2}, even though we are using similar terminology, and frame it in a similar way. In argument coreference, it is equally as important to link two coreferent argument mentions as it is not to link two different arguments. In contrast, regular coreference emphasizes the prediction of links between mentions. We thus report a different metric in Tables~\ref{tab:argument_identification}~and~\ref{tab:cascade_argument_identification}, \emph{exact match coreference}, which gives credit for returning a cluster of mentions that corresponds exactly to an argument. In Figure~\ref{fig:subtasks}, a system would be rewarded for linking together both mentions of the taxpayer in §2(a)(1)(B), but not if any of the two mentions were linked to any other mention within §2(a)(1)(B). This custom metric gives as much credit for correctly linking a single-mention argument (no links), as for a 5-mention argument (10 links).

\mysubsubsection{Single mention baseline}

Here, we predict no coreference links. Under usual coreference metrics, this system can have low performance.

\mysubsubsection{String matching baseline}

This baseline predicts a coreference link if the placeholder strings of two arguments are identical, up to the presence of the words \emph{such}, \emph{a}, \emph{an}, \emph{the}, \emph{any}, \emph{his} and \emph{every}.

\begin{table}[h!]
    \centering
    \begin{tabular}{lcc}
        \textbf{Single mention}   & avg $\pm$ stddev & macro \\
        \hline precision & 81.7 $\pm$ 28.9 & 68.2 \\
        recall           & 86.9 $\pm$ 21.8 & 82.7 \\
        F1               & 83.8 $\pm$ 26.0 & 74.8 \\
        \textbf{String matching}  & avg $\pm$ stddev & macro\\
        \hline precision & 91.2 $\pm$ 20.0 & 85.5 \\
        recall           & 92.8 $\pm$ 16.8 & 89.4 \\
        F1               & 91.8 $\pm$ 18.6 & 87.4 \\
    \end{tabular}
    \caption{Exact match coreference results. Average and standard deviations are computed across subsections.}
    \label{tab:argument_identification}
\end{table}

We also provide usual coreference metrics in Table~\ref{tab:coreference_metrics}, using the code associated with \citet{pradhan-EtAl:2014:P14-2}. This baseline perfectly resolves coreference for 80.8\% of subsections, \emph{versus} 68.9\% for the single mention baseline.

\begin{table}[h!]
    \centering
    \begin{tabular}{lcc}
                    & Single mention  & String matching    \\
        \hline MUC  & \phantom{0}0\phantom{.0} / \phantom{0}0\phantom{.0} / \phantom{0}0\phantom{.0} & 82.1 / 64.0 / 71.9  \\
        CEAF$_m$    & 82.5 / 82.5 / 82.5 & 92.1 / 92.1 / 92.1 \\
        CEAF$_e$    & 77.3 / 93.7 / 84.7 & 90.9 / 95.2 / 93.0 \\
        BLANC       & 50.0 / 50.0 / 50.0 & 89.3 / 81.0 / 84.7 \\
    \end{tabular}
    \caption{Argument coreference baselines scored with usual metrics. Results are shown as Precision~/~Recall~/~F1.}
    \label{tab:coreference_metrics}
\end{table}

In addition, we provide a cascade of the best methods for argument identification and coreference, and report results in Table~\ref{tab:cascade_argument_identification}. The cascade perfectly resolves a subsection's arguments in only 16.4\% of cases. This setting, which groups the first two tasks together, offers a significant challenge.

\begin{table}[h!]
    \centering
    \begin{tabular}{lcc}
        \textbf{Cascade}    & avg $\pm$ stddev  & macro \\
        \hline precision    & 54.5 $\pm$ 35.6   & 58.0 \\
        recall              & 53.5 $\pm$ 37.2   & 52.4 \\
        F1                  & 54.7 $\pm$ 33.4   & 55.1 \\
    \end{tabular}
    \caption{Exact match coreference results for BERT-based argument identification followed by string matching-based argument coreference. Average and standard deviations are computed across subsections.}
    \label{tab:cascade_argument_identification}
\end{table}

\subsection{Argument instantiation}
\label{subsec:argument_instantiation}

Argument instantiation takes into account the information provided by previous tasks. We start by instantiating the arguments of a single subsection, without regard to the structure of the statutes. We then describe how the structure information is incorporated into the model.

\mysubsubsection{Single subsection}
\label{sec:single_subsection}

\begin{algorithm}[h!]
\small
    \caption{Argument instantiation for a single subsection}
    \label{alg:argument_instantiation_single}
    \begin{algorithmic}[1]
        \Require argument spans with coreference information $A$, input argument-value pairs $D$, subsection text $s$, case description $c$
        \Ensure output argument-value pairs $P$
        \Function{ArgInstantiation}{$A, D, s, c$}
            \State $P \gets \emptyset$
            \For{$a$ in $A \setminus \{\texttt{@truth}\}$} \label{sfs:for_start} 
                \State $r \gets \textsc{InsertValues}(s,A,D,P)$ \label{sfs:insert_values}
                \State $y \gets \textsc{BERT}(c,r)$ \label{sfs:bert}
                \State $x \gets \textsc{ComputeAttentiveReps}(y,a)$ \label{sfs:attentive_representations}
                \State $v \gets \textsc{PredictValue}(x)$ \label{sfs:decoder}
                \State $P \gets P \cup (a,v)$ \label{sfs:insert_into_dict}
            \EndFor \label{sfs:for_stop}
            \State $r \gets \textsc{InsertValues}(s,A,D,P)$ \label{sfs:insert_values_2}
            \State $y \gets \textsc{BERT\_CLS}(c,r)$ \label{sfs:bert_cls}
            \State $t \gets \textsc{TruthPredictor}(y)$ \label{sfs:predictor}
            \State $P \gets P \cup (\texttt{@truth},t)$ \label{sfs:insert_truth_value} \\
            \Return $P$
        \EndFunction
    \end{algorithmic}
\end{algorithm}

We follow the paradigm of \citet{chen2020reading}, where we iteratively modify the text of the subsection by inserting argument values, and predict values for uninstantiated arguments. Throughout the following, we refer to Algorithm~\ref{alg:argument_instantiation_single} and to its notation.

For each argument whose value is provided, we replace the argument's placeholders in subsection $s$ by the argument's value, using \textsc{InsertValues} (line~\ref{sfs:insert_values}). This yields mostly grammatical sentences, with occasional hiccups. With §2(a)(1)(A) and the top right case from Figure~\ref{fig:subtasks}, we obtain ``(A) Alice spouse died during either of the two years immediately preceding 2017''.

We concatenate the text of the case $c$ with the modified text of the subsection $r$, and embed it using BERT (line~\ref{sfs:bert}), yielding a sequence of contextual subword embeddings ${y = \{ y_i \in \mathbb{R}^{768} \thinspace | \thinspace  i = 1 ... n \}}$. Keeping with the notation of \citet{chen2020reading}, assume that the embedded case is represented by the sequence of vectors ${\myvector{t}_1,..., \myvector{t}_m}$ and the embedded subsection by ${\myvector{s}_1,..., \myvector{s}_n}$. For a given argument $a$, compute its attentive representation ${\myvector{\tilde s}_1,..., \myvector{\tilde s}_m}$ and its augmented feature vectors ${\myvector{x}_1,..., \myvector{x}_m}$. This operation, described by \citet{chen2020reading}, is performed by \textsc{ComputeAttentiveReps} (line~\ref{sfs:attentive_representations}). The augmented feature vectors ${\myvector{x}_1,..., \myvector{x}_m}$ represent the argument's placeholder, conditioned on the text of the statute and case.

Based on the name of the argument span, we predict its value $v$ either as an integer or a span from the case description, using \textsc{PredictValue} (line~\ref{sfs:decoder}). For integers, as part of the model training, we run k-means clustering on the set of all integer values in the training set, with enough centroids such that returning the closest centroid instead of the true value yields a numerical accuracy of 1 (see below). For any argument requiring an integer (e.g. \texttt{tax}), the model returns a weighted average of the centroids. The weights are predicted by a linear layer followed by a softmax, taking as input an average-pooling and a maxpooling of $\myvector{x}_1,..., \myvector{x}_m$. For a span from the case description, we follow the standard procedure for fine-tuning BERT on SQuAD \citep{devlin2018bert}. The unnormalized probability of the span from tokens $i$ to $j$ is given by $e^{\myvector{l}\cdot \myvector{x}_i + \myvector{r}\cdot \myvector{x}_j}$ where $\myvector{l}, \myvector{r}$ are learnable parameters.

The predicted value $v$ is added to the set of predictions $P$ (line~\ref{sfs:insert_into_dict}), and will be used in subsequent iterations to replace the argument's placeholder in the subsection. We repeat this process until a value has been predicted for every argument, except \texttt{@truth} (lines~\ref{sfs:for_start}-\ref{sfs:for_stop}). Arguments are processed in order of appearance in the subsection. Finally, we concatenate the case and fully grounded subsection and embed them with BERT (lines~\ref{sfs:insert_values_2}-\ref{sfs:bert_cls}), then use a linear predictor on top of the representation for the [CLS] token to predict the value for the \texttt{@truth} argument (line~\ref{sfs:predictor}).

\mysubsubsection{Subsection with dependencies}

\begin{algorithm}[h!]
\small
    \caption{Argument instantiation with dependencies}
    \label{alg:argument_instantiation_full}
    \begin{algorithmic}[1]
        \Require argument spans with coreference information $A$, structure information $T$, input argument-value pairs $D$, subsection $s$, case description $c$
        \Ensure output argument-value pairs $P$
        \Function{ArgInstantiationFull}{$A, T, D, s, c$}
            \State $t \gets \textsc{BuildDependencyTree}(s,T)$ \label{sff:tree_start}
            \State $t \gets \textsc{PopulateArgValues}(t,D)$
            \State $Q \gets \textrm{depth-first traversal of } t$\label{sff:tree_end}
            \For{$q$ in $Q$}
                \If{$q$ is a subsection and a leaf node} \label{sff:leaf_start}
                    \State $D_q \gets \textsc{GetArgValuePairs}(q)$ 
                    \State $\tilde s \gets \textsc{GetSubsectionText}(q)$ 
                    \State $q \gets \textsc{ArgInstantiation}(A,D_q,\tilde s,c)$  \label{sff:leaf_end}
                \ElsIf{$q$ is a subsection and not a leaf node}
                    \State $D_q \gets \textsc{GetArgValuePairs}(q)$ \label{sff:argvalue_pairs}
                    \State $x \gets \textsc{GetChild}(q)$ \label{sff:get_child}
                    \State $D_x \gets \textsc{GetArgValuePairs}(x)$
                    \label{sff:argvalue_pairs_from_child}
                    \State $D_q \gets D_\mathrm{q} \cup D_x$ \label{sff:merge_argvalue_pairs}
                    \State $\tilde s \gets \textsc{GetSubsectionText}(q)$ \label{sff:subsection_text}
                    \State $q \gets \textsc{ArgInstantiation}(A,D_q,\tilde s,c)$  \label{sff:subsection_arginstantiation}
                \ElsIf{$q \in \{\textsc{AND}, \textsc{OR}, \textsc{NOT}\}$} \label{sff:operation_start}
                    \State $C \gets \textsc{GetChildren}(q)$ \label{sff:get_children}
                    \State $q \gets \textsc{DoOperation}(C,q)$ \label{sff:operation_end}
                \EndIf
            \EndFor
            \State $x \gets \textsc{Root}(t)$
            \State $P \gets \textsc{GetArgValuePairs}(x)$
            \\ \Return $P$
        \EndFunction
    \end{algorithmic}
\end{algorithm}

\begin{figure*}[h!]
    \centering
    \includegraphics[width=\textwidth]{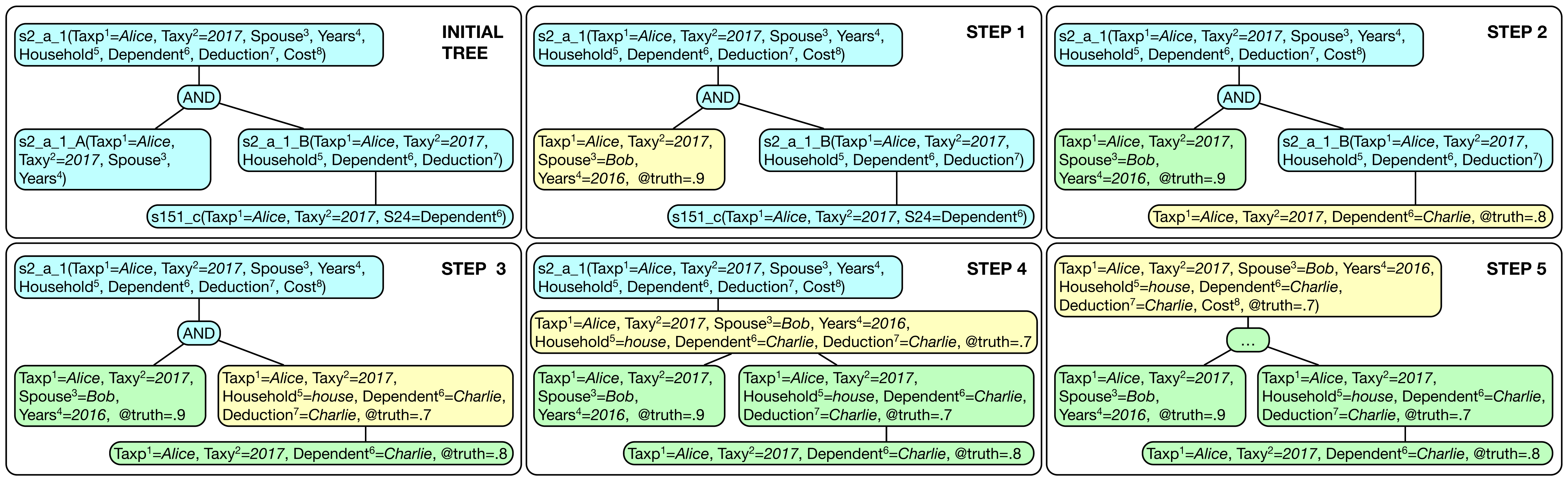}
    \caption{Argument instantiation with the top example from Figure~\ref{fig:subtasks}. At each step, nodes to be processed are in blue, nodes being processed in yellow, and nodes already processed in green. The last step was omitted, and involves determining the truth value of the root node's \texttt{@truth} argument.}
    \vspace{-1pt}
    \label{fig:argument_instantiation_tree}
\end{figure*}

To describe our procedure at a high-level, we use the structure of the statutes to build out a computational graph, where nodes are either subsections with argument-value pairs, or logical operations. We resolve nodes one by one, depth first. We treat the single-subsection model described above as a function, taking as input a set of argument-value pairs, a string representation of a subsection, and a string representation of a case, and returning a set of argument-value pairs. Algorithm~\ref{alg:argument_instantiation_full} and Figure~\ref{fig:argument_instantiation_tree} summarize the following.

We start by building out the subsection's dependency tree, as specified by the structure annotations (lines~\ref{sff:tree_start}-\ref{sff:tree_end}). First, we build the tree structure using \textsc{BuildDependencyTree}. Then, values for arguments are propagated from parent to child, from the root down, with \textsc{PopulateArgValues}. The tree is optionally capped to a predefined depth. Each node is either an input for the single-subsection function or its output, or a logical operation. We then traverse the tree depth first, performing the following operations, and replacing the node with the result of the operation:

\begin{itemize} [leftmargin=*]
    \item If the node $q$ is a leaf, resolve it using the single-subsection function \textsc{ArgInstantiation} (lines~\ref{sff:leaf_start}-\ref{sff:leaf_end} in Algorithm~\ref{alg:argument_instantiation_full}; step~1 in Figure~\ref{fig:argument_instantiation_tree}).
    
    \item If the node $q$ is a subsection that is not a leaf, find its child node $x$ (\textsc{GetChild}, line~\ref{sff:get_child}), and corresponding argument-value pairs other than \texttt{@truth}, $D_x$ (\textsc{GetArgValuePairs}, line~\ref{sff:argvalue_pairs_from_child}). Merge $D_x$ with $D_q$, the argument-value pairs of the main node $q$ (line~\ref{sff:merge_argvalue_pairs}). Finally, resolve the parent node $q$ using the single-subsection function (lines~\ref{sff:subsection_text}-\ref{sff:subsection_arginstantiation}; step~3 in Figure~\ref{fig:argument_instantiation_tree}.
    
    \item If node $q$ is a logical operation (line~\ref{sff:operation_start}), get its children $C$ (\textsc{GetChildren}, line~\ref{sff:get_children}), to which the operation will be applied with \textsc{DoOperation} (line~\ref{sff:operation_end}) as follows:
    \begin{itemize}
        \item If $q==\textsc{NOT}$, assign the negation of the child's \texttt{@truth} value to $q$.
        
        \item If $q==\textsc{OR}$, pick its child with the highest \texttt{@truth} value, and assign its arguments' values to $q$.
        
        \item If $q==\textsc{AND}$, transfer the argument-value pairs from all its children to $q$. In case of conflicting values, use the value associated with the lower \texttt{@truth} value. This operation can be seen in step~4 of Figure~\ref{fig:argument_instantiation_tree}.
    \end{itemize}
\end{itemize}

This procedure follows the formalism of neural module networks \citep{andreas2016neural} and is illustrated in Figure~\ref{fig:argument_instantiation_tree}.
Reentrancy into the dependency tree is not possible, so that a decision made earlier cannot be backtracked on at a later stage. One could imagine doing joint inference, or using heuristics for revisiting decisions, for example with a limited number of reentrancies. Humans are generally able to resolve this task in the order of the text, and we assume it should be possible for a computational model too. Our solution is meant to be computationally efficient, with the hope of not sacrificing too much performance. Revisiting this assumption is left for future work.

\mysubsubsection{Metrics and evaluation}

Arguments whose value needs to be predicted fall into three categories. The \texttt{@truth} argument calls for a binary truth value, and we score a model's output using binary accuracy. The values of some arguments, such as \texttt{gross income}, are dollar amounts. We score such values using numerical accuracy, as $1$~if~$\Delta(y,\hat y) = \frac{|y-\hat y|}{\textrm{max}(0.1*y,5000)} < 1$~else~$0$, where $\hat y$ is the prediction and $y$ the target. All other argument values are treated as strings. In those cases, we compute accuracy as exact match between predicted and gold value. Each of these three metrics defines a form of accuracy. We average the three metrics, weighted by the number of samples, to obtain a unified accuracy metric, used to compare the performance of models.

\begin{table*}
    \centering
    \begin{tabular}{p{59pt}ccccccc}
                                                        &  \small \texttt{@truth}   & \small \texttt{dollar amount} & \small \texttt{string}                & unified                   & \vline & binary                   & numerical     \\
    \hline \cellcolor[rgb]{.953 .812 .5} baseline       & 58.3 $\pm$ 7.5            & 18.2 $\pm$ 11.5               & \phantom{0}4.4 $\pm$ \phantom{0}7.4   & 43.3 $\pm$ 6.2            & \vline & 50 $\pm$ 8.3             & 30 $\pm$ 18.1 \\
    \cellcolor[rgb]{.620 .859 .957} + silver            & 58.3 $\pm$ 7.5            & \textbf{39.4} $\pm$ 14.6      & \phantom{0}4.4 $\pm$ \phantom{0}7.4   & 47.2 $\pm$ 6.2            & \vline & 50 $\pm$ 8.3             & \textbf{45} $\pm$ 19.7 \\
    \cellcolor[rgb]{.620 .859 .957} BERT                & 59.2 $\pm$ 7.5            & 23.5 $\pm$ 12.5               & \textbf{37.5} $\pm$ 17.3              & 49.4 $\pm$ 6.2            & \vline & 51 $\pm$ 8.3             & 30 $\pm$ 18.1 \\
    \cellcolor[rgb]{.953 .812 .5} - pre-training        & 57.5 $\pm$ 7.5            & 20.6 $\pm$ 11.9               & \textbf{37.5} $\pm$ 17.3              & 47.8 $\pm$ 6.2            & \vline & 49 $\pm$ 8.3             & 30 $\pm$ 18.1 \\
    \cellcolor[rgb]{.620 .859 .957} - structure         & \textbf{65.8} $\pm$ 7.2   & 20.6 $\pm$ 11.9               & 33.3 $\pm$ 16.8                       & \textbf{52.8} $\pm$ 6.2   & \vline & \textbf{59} $\pm$ 8.2    & 30 $\pm$ 18.1 \\
    \cellcolor[rgb]{.953 .812 .5} - pre-training,       & 60.8 $\pm$ 7.4            & 20.6 $\pm$ 11.9               & 33.3 $\pm$ 16.8                       & 49.4 $\pm$ 6.2            & \vline & 53 $\pm$ 8.3             & 30 $\pm$ 18.1 \\
    \cellcolor[rgb]{.953 .812 .5} \phantom{-} structure &                           &                               &                                       &                   &                       & \multicolumn{2}{r}{\small{(best results in \textbf{bold})}} \\
    \end{tabular}
    \caption{Argument instantiation. We report accuracies, in \%, and the 90\% confidence interval. Right of the bar are accuracy metrics proposed with the initial release of the dataset. Blue cells use the silver data, brown cells do not. ``BERT'' is the model described in Section~\ref{subsec:argument_instantiation}. Ablations to it are marked with a ``-'' sign.}
    \vspace{-1pt}
    \label{tab:argument_instantiation_results}
\end{table*}

\mysubsubsection{Training}

Based on the type of value expected, we use different loss functions. For \texttt{@truth}, we use binary cross-entropy. For numerical values, we use the hinge loss $\textrm{max}(\Delta(y,\hat y)-1,0)$. For strings, let $S$ be all the spans in the case description equal to the expected value. The loss function is $\log(\sum_{i\leq j} e^{\myvector{l}\cdot \myvector{x}_i + \myvector{r}\cdot \myvector{x}_j}) -\log(\sum_{i,j \in S} e^{\myvector{l}\cdot \myvector{x}_i + \myvector{r}\cdot \myvector{x}_j})$ \citep{clark2018simple}. The model is trained end-to-end with gradient descent.

We start by training models on the silver data, as a pre-training step. We sweep the values of the learning rate in $\{10^{-2}, 10^{-3}, 10^{-4}, 10^{-5}\}$ and the batch size in $\{64, 128, 256\}$. We try both BERT-base-cased and Legal BERT, allowing updates to the parameters of its top layer. We set aside 10\% of the silver data as a dev set, and select the best model based on the unified accuracy on the dev set. Training is split up into three stages. The single-subsection model iteratively inserts values for arguments into the text of the subsection. In the first stage, regardless of the predicted value, we insert the gold value for the argument, as in teacher forcing \citep{kolen2001field}. In the second and third stages, we insert the value predicted by the model. When initializing the model from one stage to the next, we pick the model with the highest unified accuracy on the dev set. In the first two stages, we ignore the structure of the statutes, which effectively caps the depth of each dependency tree at 1.

Picking the best model from this pre-training step, we perform fine-tuning on the gold data. We take a k-fold cross-validation approach \citep{stone1974cross}. We randomly split the SARA v2 training set into 10 splits, taking care to put pairs of cases testing the same subsection into the same split. Each split contains nearly exactly the same proportion of binary and numerical cases. We sweep the values of the learning rate and batch size in the same ranges as above, and optionally allow updates to the parameters of BERT's top layer. For a given set of hyperparameters, we run training on each split, using the dev set and the unified metric for early stopping. We use the performance on the dev set averaged across the 10 splits to evaluate the performance of a given set of hyperparameters. Using that criterion, we pick the best set of hyperparameters. We then pick the final model as that which achieves median performance on the dev set, across the 10 splits. We report the performance of that model on the test set.

In Table~\ref{tab:argument_instantiation_results}, we report the relevant argument instantiation metrics, under \texttt{@truth}, \texttt{dollar amount} and \texttt{string}. For comparison, we also report binary and numerical accuracy metrics defined in \citet{holzenberger2020dataset}. The reported baseline has three parameters. For \texttt{@truth}, it returns the most common value for that argument on the train set. For arguments that call for a dollar amount, it returns the one number that minimizes the \texttt{dollar amount} hinge loss on the training set. For all other arguments, it returns the most common string answer in the training set. Those parameters vary depending on whether the training set is augmented with the silver data.

\section{Discussion}
\label{sec:discussion}

Our goal in providing the baselines of Section~\ref{sec:baseline} is to identify performance bottlenecks in the proposed sequence of tasks. Argument identification poses a moderate challenge, with a language model-based approach achieving non-trivial F1 score. The simple parser-based method is not a sufficient solution, but with its high recall could serve as the backbone to a statistical method. Argument coreference is a simpler task, with string matching perfectly resolving nearly 80\% of the subsections. This is in line with the intuition that legal language is very explicit about disambiguating coreference. As reported in Table~\ref{tab:coreference_metrics}, usual coreference metrics seem lower, but only reflect a subset of the full task: coreference metrics are only concerned with links, so that arguments appearing exactly once bear no weight under that metric, unless they are wrongly linked to another argument.

Argument instantiation is by far the most challenging task, as the model needs strong natural language understanding capabilities. Simple baselines can achieve accuracies above 50\% for \texttt{@truth}, since for all numerical cases, \texttt{@truth}~=~True. We receive a slight boost in binary accuracy from using the proposed paradigm, departing from previous results on this benchmark. As compared to the baseline, the models mostly lag behind for the \texttt{dollar amount} and numerical accuracies, which can be explained by the lack of a dedicated numerical solver, and sparse data. Further, we have made a number of simplifying assumptions, which may be keeping the model from taking advantage of the structure information: arguments are instantiated in order of appearance, forbidding joint prediction; revisiting past predictions is disallowed, forcing the model to commit to wrong decisions made earlier; the depth of the dependency tree is capped at 3; and finally, information is being passed along the dependency tree in the form of argument values, as opposed to dense, high-dimensional vector representations. The latter limits both the flow of information and the learning signal. This could also explain why the use of dependencies is detrimental in some cases. Future work would involve joint prediction \citep{chan2019kermit}, and more careful use of structure information.

Looking at the errors made by the best model in Table~\ref{tab:argument_instantiation_results} for binary accuracy, we note that for 39 positive and negative case pairs, it answers each pair identically, thus yielding 39 correct answers. In the remaining 11 pairs, there are 10 pairs where it gets both cases right. This suggests it may be guessing randomly on 39 pairs, and understanding 10. The best BERT-based model for \texttt{dollar amounts} predicts the same number for each case, as does the baseline. The best models for \texttt{string} arguments generally make predictions that match the category of the expected answer (date, person, etc) while failing to predict the correct string.

Performance gains from silver data are noticeable and generally consistent, as can be seen by comparing brown and blue cells in Table~\ref{tab:argument_instantiation_results}. The silver data came from running a human-written Prolog program, which is costly to produce. A possible substitute is to find mentions of applicable statutes in large corpora of legal cases \citep{caselawproject}, for example using high-precision rules \citep{ratner2017snorkel}, which has been successful for extracting information from cases \citep{boniol2020performance}.

In this work, each task uses the gold annotations from upstream tasks. Ultimately, the goal is to pass the outputs of models from one task to the next.

\section{Related Work}
\label{sec:related_work}

Law-related NLP tasks have flourished in the past years, with applications including answering bar exam questions \citep{yoshioka2018overview,zhong2019jec}, information extraction \citep{chalkidis2019large, boniol2020performance, lam2020gap}, managing contracts \citep{elwany2019bert,liepina2020explaining,nyarko2021stickiness} and analyzing court decisions \citep{sim2014utility,lee2017judging}. Case-based reasoning has been approached with expert systems \citep{popp1974judith,hellawell1980computer,vdl1983design}, high-level hand-annotated features \citep{ashley2009automatically} and transformer-based models \citep{rabelo2019combining}. Closest to our work is \citet{saeidi2018interpretation}, where a dialog agent's task is to answer a user's question about a set of regulations. The task relies on a set of questions provided within the dataset.

\citet{clark2019f} as well as preceding work \citep{friedland2004project,gunning2010project} tackle a similar problem in the science domain, with the goal of using the prescriptive knowledge from science textbooks to answer exam questions. The core of their model relies on several NLP and specialized reasoning techniques, with contextualized language models playing a major role. \citet{clark2019f} take the route of sorting questions into different types, and working on specialized solvers. In contrast, our approach is to treat each question identically, but to decompose the process of answering into a sequence of subtasks.

The language of statutes is related to procedural language, which describes steps in a process. \citet{zhang2012automatically} collect how-to instructions in a variety of domains, while \citet{wambsganss2019mining} focus on automotive repair instructions. \citet{branavan-etal-2012-learning} exploit instructions in a game manual to improve an agent's performance. \citet{dalvi-etal-2019-everything} and \citet{amini2020procedural} turn to modeling textual descriptions of physical and biological mechanisms. \citet{weller2020learning} propose models that generalize to new task descriptions.

The tasks proposed in this work are germane to standard NLP tasks, such as named entity recognition \citep{ratinov2009design}, part-of-speech tagging \citep{petrov2011universal,akbik2018contextual}, and coreference resolution \citep{pradhan-EtAl:2014:P14-2}. Structure extraction is conceptually similar to syntactic \citep{socher2013parsing} and semantic parsing \citep{berant2013semantic}, which \citet{pertierra2017towards} attempt for a subsection of tax law.

Argument instantiation is closest to the task of aligning predicate argument structures \citep{roth2012aligning,wolfe-etal-2013-parma}. We frame argument instantiation as iteratively completing a statement in natural language. \citet{chen2020reading} refine generic statements by copying strings from input text, with the goal of detecting events. \citet{chan2019kermit} extend transformer-based language models to permit inserting tokens anywhere in a sequence, thus allowing to modify an existing sequence. For argument instantiation, we make use of neural module networks \citep{andreas2016neural}, which are used in the visual \cite{yi2018neural} and textual domains \citep{gupta2019neural}. In that context, arguments and their values can be thought of as the hints from \citet{khot2020text}. The Prolog-based data augmentation is related to data augmentation for semantic parsing \citep{campagna2019genie,weir2019dbpal}.

\section{Conclusion}

Solutions to tackle statutory reasoning may range from high-structure, high-human involvement expert systems, to less structured, largely self-supervised language models. Here, taking inspiration from Prolog programs, we introduce a novel paradigm, by breaking statutory reasoning down into a sequence of tasks. Each task can be annotated for with far less expertise than would be required to translate legal language into code, and comes with its own performance metrics. Our contribution enables finer-grained scoring and debugging of models for statutory reasoning, which facilitates incremental progress and identification of performance bottlenecks. In addition, argument instantiation and explicit resolution of dependencies introduce further interpretability. This novel approach could possibly inform the design of models that reason with rules specified in natural language, for the domain of legal NLP and beyond.

\section*{Acknowledgments}

The authors thank Andrew Blair-Stanek for helpful comments, and Ryan Culkin for help with the parser-based argument identification baseline.

\clearpage

\bibliographystyle{acl_natbib}
\bibliography{anthology,acl2021}

\clearpage

\appendix

\section{Task examples}
\label{app:task_examples}

In the following, we provide several examples for each of the tasks defined in Section \ref{sec:sara}.

\subsection{Argument identification}

For ease of reading, the spans mentioned in the output are \underline{underlined} in the input.

\myexample{Input 1 (§3306(a)(1)(B))}

\noindent (B) on each of \underline{some 10 days} during \underline{the calendar year} or during \underline{the preceding calendar year}, \underline{each day} being in a different \underline{calendar week}, employed at least \underline{one individual} in \underline{employment} for \underline{some portion of the day}.

\myexample{Output 1}

\noindent $\{(15,26)\mymathcomma(35, 51)\mymathcomma(62, 88)\mymathcomma(92, 99)\mymathcomma(122, 134)\mymathcomma(155, 168)\mymathcomma(173, 182)\mymathcomma(188, 210)\}$

\myexample{Input 2 (§63(c)(5))}

\noindent In the case of \underline{an individual} with respect to whom \underline{a deduction} under section 151 is allowable to \underline{another taxpayer} for \underline{a taxable year} beginning in \underline{the calendar year} in which \underline{the individual}'s \underline{taxable year} begins, \underline{the basic standard deduction} applicable to \underline{such individual} for \underline{such individual}'s \underline{taxable year} shall not exceed \underline{the greater} of-

\myexample{Output 2}

\noindent $\{(15, 27)\mymathcomma(50, 60)\mymathcomma(96, 111)\mymathcomma(117, 130)\mymathcomma(145, 161)\mymathcomma(172, 185)\mymathcomma (189, 200)\mymathcomma(210, 237)\mymathcomma(253, 267)\mymathcomma(273, 287)\mymathcomma(291, 302)\mymathcomma(321, 331)\}$

\myexample{Input 3 (§1(d)(iv))}

\noindent (iv) \underline{\$31,172, plus 36\% of the excess over \$115,000} if \underline{the taxable income} is over \$115,000 but not over \$250,000;

\myexample{Output 3}

\noindent $\{(5,45)\mymathcomma(50, 67)\}$

\subsection{Argument coreference}

We report the full matrix $C$. In addition, for ease of reading, coreference clusters are marked with superscripts in the input.

\myexample{Input 1 (§3306(a)(1)(B))}

\noindent (B) on each of \underline{some 10 days}${}^1$  during \underline{the calendar year}${}^2$ or during \underline{the preceding calendar year}${}^3$, \underline{each day}${}^1$  being in a different \underline{calendar week}${}^4$, employed at least \underline{one individual}${}^5$ in \underline{employment}${}^6$ for \underline{some portion of the day}${}^7$.

\noindent $\{(15,26)\mymathcomma(35, 51)\mymathcomma(62, 88)\mymathcomma(92, 99)\mymathcomma(122, 134)\mymathcomma(155, 168)\mymathcomma(173, 182)\mymathcomma(188, 210)\}$

\myexample{Output 1}

$$\left( \begin{array}{cccccccc}
1&0&0&1&0&0&0&0 \\
0&1&0&0&0&0&0&0 \\
0&0&1&0&0&0&0&0 \\
1&0&0&1&0&0&0&0 \\
0&0&0&0&1&0&0&0 \\
0&0&0&0&0&1&0&0 \\
0&0&0&0&0&0&1&0 \\
0&0&0&0&0&0&0&1 \\
\end{array}\right)$$

\myexample{Input 2 (§63(c)(5))}

\noindent In the case of \underline{an individual}${}^1$ with respect to whom \underline{a deduction}${}^2$ under section 151 is allowable to \underline{another taxpayer}${}^3$ for \underline{a taxable year}${}^4$ beginning in \underline{the calendar year}${}^5$ in which \underline{the individual}${}^1$'s \underline{taxable year}${}^6$ begins, \underline{the basic standard deduction}${}^7$ applicable to \underline{such individual}${}^1$ for \underline{such individual}${}^1$'s \underline{taxable year}${}^6$ shall not exceed \underline{the greater}${}^8$ of-

\noindent $\{(15, 27)\mymathcomma(50, 60)\mymathcomma(96, 111)\mymathcomma(117, 130)\mymathcomma(145, 161)\mymathcomma(172, 185)\mymathcomma (189, 200)\mymathcomma(210, 237)\mymathcomma(253, 267)\mymathcomma(273, 287)\mymathcomma(291, 302)\mymathcomma(321, 331)\}$

\myexample{Output 2}

$$\left( \begin{array}{cccccccccccc}
1&0&0&0&0&1&0&0&1&1&0&0 \\
0&1&0&0&0&0&0&0&0&0&0&0 \\
0&0&1&0&0&0&0&0&0&0&0&0 \\
0&0&0&1&0&0&0&0&0&0&0&0 \\
0&0&0&0&1&0&0&0&0&0&0&0 \\
1&0&0&0&0&1&0&0&1&1&0&0 \\
0&0&0&0&0&0&1&0&0&0&1&0 \\
0&0&0&0&0&0&0&1&0&0&0&0 \\
1&0&0&0&0&1&0&0&1&1&0&0 \\
1&0&0&0&0&1&0&0&1&1&0&0 \\
0&0&0&0&0&0&1&0&0&0&1&0 \\
0&0&0&0&0&0&0&0&0&0&0&1 \\
\end{array}\right)$$

\myexample{Input 3 (§1(d)(iv))}

\noindent (iv) \underline{\$31,172, plus 36\% of the excess over \$115,000}${}^1$ if \underline{the taxable income}${}^2$ is over \$115,000 but not over \$250,000;

\noindent $\{(5,45)\mymathcomma(50, 67)\}$

\clearpage

\myexample{Output 3}

$$\left( \begin{array}{cc}
1 & 0 \\
0 & 1 \\
\end{array}\right)$$

\subsection{Structure extraction}

To clarify the link between the input and the output, we are adding superscripts to argument names in the output. While the output is represented as plain text, a graph-based representation would likely be used in a practical system, to facilitate learning and inference. Arguments are keyword based. For example, in Output 2, the value of the Taxp argument of §63(c)(5) is passed to the Spouse argument of §151(b). If no equal sign is specified, it means the argument names match. For example, part of Output 2 could have been rewritten more explicitly as §151(b)(Spouse=Taxp, Taxp=S45, Taxy=Taxy).

\myexample{Input 1 (§3306(a)(1)(B))}

\noindent (B) on each of \underline{some 10 days}${}^1$  during \underline{the calendar year}${}^2$ or during \underline{the preceding calendar year}${}^3$, \underline{each day}${}^1$  being in a different \underline{calendar week}${}^4$, employed at least \underline{one individual}${}^5$ in \underline{employment}${}^6$ for \underline{some portion of the day}${}^7$.

\noindent $\{(15,26)\mymathcomma(35, 51)\mymathcomma(62, 88)\mymathcomma(92, 99)\mymathcomma(122, 134)\mymathcomma(155, 168)\mymathcomma(173, 182)\mymathcomma(188, 210)\}$

$$\left( \begin{array}{cccccccc}
1&0&0&1&0&0&0&0 \\
0&1&0&0&0&0&0&0 \\
0&0&1&0&0&0&0&0 \\
1&0&0&1&0&0&0&0 \\
0&0&0&0&1&0&0&0 \\
0&0&0&0&0&1&0&0 \\
0&0&0&0&0&0&1&0 \\
0&0&0&0&0&0&0&1 \\
\end{array}\right)$$

\needspace{3\baselineskip}
\myexample{Output 1}

\begin{small}
    \begin{tabbing}
        §3306(a)(1)(B)(\= Caly${}^2$, S16${}^7$, Workday${}^1$, Employment${}^6$, \\
        \> Preccaly${}^3$, Employee${}^5$, S13A${}^4$, Employer, \\
        \> Service) :- \\
        \indent §3306(c)(Employee, Employer, Service).
    \end{tabbing}
\end{small}

\myexample{Input 2 (§63(c)(5))}

\noindent In the case of \underline{an individual}${}^1$ with respect to whom \underline{a deduction}${}^2$ under section 151 is allowable to \underline{another taxpayer}${}^3$ for \underline{a taxable year}${}^4$ beginning in \underline{the calendar year}${}^5$ in which \underline{the individual}${}^1$'s \underline{taxable year}${}^6$ begins, \underline{the basic standard deduction}${}^7$ applicable to \underline{such individual}${}^1$ for \underline{such individual}${}^1$'s \underline{taxable year}${}^6$ shall not exceed \underline{the greater}${}^8$ of-

\noindent $\{(15, 27)\mymathcomma(50, 60)\mymathcomma(96, 111)\mymathcomma(117, 130)\mymathcomma(145, 161)\mymathcomma(172, 185)\mymathcomma (189, 200)\mymathcomma(210, 237)\mymathcomma(253, 267)\mymathcomma(273, 287)\mymathcomma(291, 302)\mymathcomma(321, 331)\}$

$$\left( \begin{array}{cccccccccccc}
1&0&0&0&0&1&0&0&1&1&0&0 \\
0&1&0&0&0&0&0&0&0&0&0&0 \\
0&0&1&0&0&0&0&0&0&0&0&0 \\
0&0&0&1&0&0&0&0&0&0&0&0 \\
0&0&0&0&1&0&0&0&0&0&0&0 \\
1&0&0&0&0&1&0&0&1&1&0&0 \\
0&0&0&0&0&0&1&0&0&0&1&0 \\
0&0&0&0&0&0&0&1&0&0&0&0 \\
1&0&0&0&0&1&0&0&1&1&0&0 \\
1&0&0&0&0&1&0&0&1&1&0&0 \\
0&0&0&0&0&0&1&0&0&0&1&0 \\
0&0&0&0&0&0&0&0&0&0&0&1 \\
\end{array}\right)$$

\needspace{8\baselineskip}
\myexample{Output 2}

\begin{small}
\begin{tabbing}
\noindent §63(c)(5)(\= Bassd${}^7$, Grossinc, S45${}^3$, Taxp${}^1$, Taxy${}^6$, \\
                        \> S44B${}^2$, S46B${}^4$, S47${}^5$, S48${}^8$) :- \\
                        \indent \= [ \\
                        \> \indent \= §151(b)(Spouse=Taxp, Taxp=S45, Taxy) OR \\
                        \> \> §151(c)(S24A=Taxp, Taxp=S45, Taxy) \\
                        \> ] AND \\
                        \> §63(c)(5)(A)() AND \\
                        \> §63(c)(5)(B)(Grossinc, Taxp).
\end{tabbing}
\end{small}

\myexample{Input 3 (§1(d)(iv))}

\noindent (iv) \underline{\$31,172, plus 36\% of the excess over \$115,000}${}^1$ if \underline{the taxable income}${}^2$ is over \$115,000 but not over \$250,000;

\noindent $\{(5,45)\mymathcomma(50, 67)\}$

$$\left( \begin{array}{cc}
1 & 0 \\
0 & 1 \\
\end{array}\right)$$

\myexample{Output 3}

\begin{small}
\noindent §1(d)(iv)(Tax${}^1$, Taxinc${}^2$).
\end{small}

\subsection{Argument instantiation}

The following are example cases. In addition to the case description,  subsection to apply and input argument-value pairs, the agent has access to the output of Argument identification, Argument coreference and Structure extraction, for the entirety of the statutes.

\myexample{Input 1: case 3306(a)(1)(B)-positive}

\noindent \emph{Case description:} Alice has employed Bob on various occasions during the year 2017: Jan 24, Feb 4, Mar 3, Mar 19, Apr 2, May 9, Oct 15, Oct 25, Nov 8, Nov 22, Dec 1, Dec 3.

\noindent \emph{Subsection to apply:} §3306(a)(1)(B)

\noindent \emph{Argument-value pairs:} \{Employer=``Alice'', Caly=``2017''\}

\myexample{Output 1}

\noindent \{Workday=[``Jan 24'', ``Feb 4'', ``Mar 3'', ``Mar 19'', ``Apr 2'', ``May 9'', ``Oct 15'', ``Oct 25'', ``Nov 8'', ``Nov 22'', ``Dec 1'', ``Dec 3''], Employee=``Bob'', Employment=``has employed'', ``S13A'': [4, 5, 9, 11, 13, 19, 41, 43, 45, 47], @truth=True\}

\myexample{Input 2: case §63(c)(5)-negative}

\noindent \emph{Case description:} In 2017, Alice was paid \$33200. Alice and Bob have been married since Feb 3rd, 2017. Bob earned \$10 in 2017. Alice and Bob file separate returns. Alice is not entitled to a deduction for Bob under section 151.

\noindent \emph{Subsection to apply:} §63(c)(5)

\noindent \emph{Argument-value pairs:} \{Taxp=``Bob'', Taxy=``2017'', Bassd=500\}

\myexample{Output 2}

\noindent \{@truth=False\}

\myexample{Input 3: tax case 5}

\noindent \emph{Case description:} In 2017, Alice's gross income was \$326332. Alice and Bob have been married since Feb 3rd, 2017, and have had the same principal place of abode since 2015. Alice was born March 2nd, 1950 and Bob was born March 3rd, 1955. Alice and Bob file separately in 2017. Bob has no gross income that year. Alice takes the standard deduction.

\noindent \emph{Subsection to apply:} Tax

\noindent \emph{Argument-value pairs:} \{Taxy=``2017'', Taxp=``Alice''\}

\myexample{Output 3}

\noindent \{Tax=116066, @truth=True\}

\clearpage

\section{Dataset statistics}
\label{app:dataset_statistics}

\subsection{Argument identification}

In Table~\ref{tab:argument_identification_statistics}, we report statistics on the annotations for the argument identification task. The numbers in that table were used to plot the top histogram in Figure~\ref{fig:stats}(a). 

\begin{table}[H]
    \centering
    \begin{tabular}{lrr}
    Counts                      & SARA          & Random        \\
    \hline \phantom{0}0         & \phantom{0}33 & \phantom{0}24 \\
    \phantom{0}1                & \phantom{0}39 & \phantom{0}16 \\
    \phantom{0}2                & \phantom{0}34 & \phantom{0}21 \\
    \phantom{0}3                & \phantom{0}32 & \phantom{0}21 \\
    \phantom{0}4                & \phantom{0}13 & \phantom{0}11 \\
    \phantom{0}5                & \phantom{0}11 & \phantom{0}12 \\
    \phantom{0}6                & \phantom{00}7 & \phantom{0}13 \\
    \phantom{0}7                & \phantom{00}4 & \phantom{00}6 \\
    \phantom{0}8                & \phantom{0}10 & \phantom{00}4 \\
    \phantom{0}9                & \phantom{00}5 & \phantom{00}1 \\
    10                          & \phantom{00}2 & \phantom{00}7 \\
    11                          & \phantom{00}2 & \phantom{00}3 \\
    12                          & \phantom{00}1 & \phantom{00}3 \\
    13                          & \phantom{00}0 & \phantom{00}2 \\
    14                          & \phantom{00}1 & \phantom{00}0 \\
    15                          & \phantom{00}0 & \phantom{00}0 \\
    16                          & \phantom{00}0 & \phantom{00}2 \\
    total                       & 194           & 146           \\
    \rule{0pt}{3ex}Statistics   &               &               \\
    \hline average              & 3.0           & 4.0           \\
    stddev                      & 2.8           & 3.6           \\
    median                      & 2\phantom{.0} & 3\phantom{.0} \\
    \end{tabular}
    \caption{Number of argument placeholders per subsection. ``Counts'' reports the number of subsections (right columns) containing a specific number of placeholders (left column). ``Random'' refers to 9 sections drawn at random from the Tax Code, and annotated.}
    \label{tab:argument_identification_statistics}
\end{table}

\subsection{Argument coreference}

In Tables \ref{tab:argument_coreference_statistics_1} and \ref{tab:argument_coreference_statistics_2}, we report statistics on the annotations for the argument coreference task. The numbers in Table~\ref{tab:argument_coreference_statistics_1} (resp.~\ref{tab:argument_coreference_statistics_2}) were used to plot the middle (resp. bottom) histogram in Figure~\ref{fig:stats}(a).

\begin{table}[H]
    \centering
    \begin{tabular}{lrr}
    Counts                      & SARA          & Random        \\
    \hline \phantom{0}0         & \phantom{0}33 & \phantom{0}24 \\
    \phantom{0}1                & \phantom{0}40 & \phantom{0}22 \\
    \phantom{0}2                & \phantom{0}44 & \phantom{0}18 \\
    \phantom{0}3                & \phantom{0}30 & \phantom{0}23 \\
    \phantom{0}4                & \phantom{0}15 & \phantom{0}18 \\
    \phantom{0}5                & \phantom{0}10 & \phantom{0}14 \\
    \phantom{0}6                & \phantom{0}13 & \phantom{0}13 \\
    \phantom{0}7                & \phantom{00}5 & \phantom{00}8 \\
    \phantom{0}8                & \phantom{00}4 & \phantom{00}2 \\
    \phantom{0}9                & \phantom{00}0 & \phantom{00}3 \\
    10                          & \phantom{00}0 & \phantom{00}1     \\
    total                       & 161           & 146               \\
    \rule{0pt}{3ex}Statistics   &               &                   \\
    \hline average              & 2.4           & 3.1               \\
    stddev                      & 2.0           & 2.4               \\
    median                      & 2\phantom{.0} & 3\phantom{.00}    \\
    \end{tabular}
    \caption{Number of arguments per subsection. ``Counts'' reports the number of subsections (right columns) containing a specific number of arguments (left column). ``Random'' refers to 9 sections drawn at random from the Tax Code, and annotated.}
    \label{tab:argument_coreference_statistics_1}
\end{table}

\begin{table}[H]
    \centering
    \begin{tabular}{lrr}
    Counts                      & SARA          & Random        \\
    \hline 1                    & \phantom{}391 & \phantom{}360 \\
    2                           & \phantom{0}70 & \phantom{0}73 \\
    3                           & \phantom{00}6 & \phantom{0}16 \\
    4                           & \phantom{00}6 & \phantom{00}6 \\
    5                           & \phantom{00}0 & \phantom{00}1 \\
    total                       & 473           & 456           \\
    \rule{0pt}{3ex}Statistics   &               &               \\
    \hline average              & 1.2           & 1.3           \\
    stddev                      & 0.5           & 0.6           \\
    median                      & 1\phantom{.0} & 1\phantom{.0} \\
    \end{tabular}
    \caption{Number of mentions per argument. ``Counts'' reports the number of arguments (right columns) mentioned a specific number of times (left column). ``Random'' refers to 9 sections drawn at random from the Tax Code, and annotated.}
    \label{tab:argument_coreference_statistics_2}
\end{table}

\subsection{Structure identification}

Table~\ref{tab:structure_extraction_statistics} reports statistics on the annotations for the structure extraction task. These numbers for arguments differ from those in Table~\ref{tab:argument_identification_statistics}, because any subsection is allowed to contain the arguments of any subsections it refers to. 

\begin{table}[H]
    \centering
    \begin{tabular}{lrr}
    Counts                      & Arguments     & Dependencies  \\
    \hline \phantom{0}0         & \phantom{00}9 & \phantom{0}80 \\
    \phantom{0}1                & \phantom{0}13 & \phantom{0}42 \\
    \phantom{0}2                & \phantom{0}40 & \phantom{0}28 \\
    \phantom{0}3                & \phantom{0}60 & \phantom{0}18 \\
    \phantom{0}4                & \phantom{0}24 & \phantom{00}8 \\
    \phantom{0}5                & \phantom{0}13 & \phantom{00}2 \\
    \phantom{0}6                & \phantom{0}14 & \phantom{00}3 \\
    \phantom{0}7                & \phantom{00}7 & \phantom{00}7 \\
    \phantom{0}8                & \phantom{00}7 & \phantom{00}1 \\
    \phantom{0}9                & \phantom{00}5 & \phantom{00}1 \\
    10                          & -             & \phantom{00}0 \\
    11                          & -             & \phantom{00}0 \\
    12                          & -             & \phantom{00}2 \\
    total                       & 192           & 192           \\
    \rule{0pt}{3ex}Statistics   &               &               \\
    \hline average              & 3.0           & 1.0           \\
    stddev                      & 2.6           & 2.4           \\
    median                      & 3\phantom{.0} & 1\phantom{.0} \\
    \end{tabular}
    \caption{Number of arguments and dependencies of each subsection, as represented in the structure annotations. ``Counts'' reports the number of arguments (right column) mentioned a specific number of times (left column).}
    \label{tab:structure_extraction_statistics}
\end{table}

\subsection{Argument instantiation}

Tables \ref{tab:argument_instantiation_input_statistics} and \ref{tab:argument_instantiation_output_statistics} show statistics for the annotations for the argument instantiation task. In the gold data, we separate training and test data, to show that both distributions are close.

\begin{table}[H]
    \centering
    \begin{tabular}{lrrrr}
                                & \multicolumn{3}{c}{Gold}                      & Silver            \\
    Counts                      & train         & test          & all           &                   \\
    \hline 0                    & \phantom{00}7 & \phantom{00}8 & \phantom{0}15 & \phantom{0}1197   \\
    1                           & \phantom{0}24 & \phantom{0}13 & \phantom{0}37 & \phantom{0}5487   \\
    2                           & 177           & \phantom{0}73 & 250           & 35629             \\
    3                           & \phantom{0}41 & \phantom{0}24 & \phantom{0}65 & 32751             \\
    4                           & \phantom{00}5 & \phantom{00}2 & \phantom{00}7 & \phantom{00}447   \\
    5                           & \phantom{00}2 & \phantom{00}0 & \phantom{00}2 & \phantom{000}32   \\
    total                       & 256           & 120           & 376           & 75543             \\
    \rule{0pt}{3ex}Statistics   &               &               &               &                   \\
    \hline average              & 2.1           & 2.0           & 2.0           & 2.3               \\
    stddev                      & 0.7           & 0.8           & 0.7           & 0.7               \\
    median                      & 2\phantom{.0} & 1\phantom{.0} & 2\phantom{.0} & 2\phantom{.0}     \\
    \end{tabular}
    \caption{Number of arguments-value pairs for the input to the argument instantiation task. ``Counts'' reports the number of arguments (right columns) mentioned a specific number of times (left column). ``Gold'' refers to the manually annotated data, and ``Silver'' to the data produced automatically through the Prolog program.}
    \label{tab:argument_instantiation_input_statistics}
\end{table}

\begin{table}[H]
    \centering
    \begin{tabular}{lrrrr}
                                & \multicolumn{3}{c}{Gold}                      & Silver            \\
    Counts                      & train         & test          & all           &                   \\
    \hline 1                    & 131           & \phantom{0}78 & 209           & 41248             \\
    2                           & \phantom{0}96 & \phantom{0}33 & 129           & 17051             \\
    3                           & \phantom{0}12 & \phantom{00}4 & \phantom{0}16 & \phantom{0}8712   \\
    4                           & \phantom{00}7 & \phantom{00}3 & \phantom{0}10 & \phantom{0}6656   \\
    5                           & \phantom{00}8 & \phantom{00}2 & \phantom{0}10 & \phantom{0}1573   \\
    6                           & \phantom{00}1 & \phantom{00}0 & \phantom{00}1 & \phantom{00}242   \\
    7                           & \phantom{00}1 & \phantom{00}0 & \phantom{00}1 & \phantom{000}51   \\
    8                           & \phantom{00}0 & \phantom{00}0 &               & \phantom{0000}8   \\
    9                           & \phantom{00}0 & \phantom{00}0 &               & \phantom{0000}2   \\
    total                       & 256           & 120           & 376           & 75543             \\
    \rule{0pt}{3ex}Statistics   &               &               &               &                   \\
    \hline average              & 1.7           & 1.5           & 1.6           & 1.8               \\
    stddev                      & 1.0           & 0.8           & 1.0           & 1.1               \\
    median                      & 1\phantom{.0} & 1\phantom{.0} & 1\phantom{.0} & 1\phantom{.0}     \\
    \end{tabular}
    \caption{Number of arguments-value pairs for the output to the argument instantiation task. ``Counts'' reports the number of arguments (right columns) mentioned a specific number of times (left column). ``Gold'' refers to the manually annotated data, and ``Silver'' to the data produced automatically through the Prolog program.}
    \label{tab:argument_instantiation_output_statistics}
\end{table}

\end{document}